\documentclass{svproc}
\usepackage{graphicx}
\usepackage{multicol}
\usepackage[bottom]{footmisc}

\usepackage[hidelinks]{hyperref}
\usepackage{placeins}
\usepackage{hhline}
\usepackage{booktabs}                                               % good-looking tables
\usepackage{graphicx}  
\usepackage{multirow}                                               % for tables
\usepackage{hyphenat}
\usepackage{subcaption}
\usepackage{amsmath}
\usepackage{url}

\begin{document}
\mainmatter              % start of a contribution
\title{Evaluating the Effectiveness of Pre-trained Language Models in Predicting the Helpfulness of Online Product Reviews}
\titlerunning{Reviews Helpfulness Prediction}  % abbreviated title (for running head)
%                                     also used for the TOC unless
%                                     \toctitle is used
%
% \author{Author 1\inst{1} \and Author 2\inst{1}}
\author{Ali Boluki \and Javad Pourmostafa Roshan Sharami \and Dimitar Shterionov}
%
%\authorrunning{Boluki et al.} % abbreviated author list (for running head)
%
\authorrunning{Boluki et al.}

\institute{Tilburg University, Tilburg 5037 AB, Netherlands\\
\email{a.boluki@tilburguniversity.edu, J.Pourmostafa@tilburguniversity.edu, D.Shterionov@tilburguniversity.edu}
}

\maketitle              % typeset the title of the contribution

\begin{abstract}

Businesses and customers can gain valuable information from product reviews. The sheer number of reviews often necessitates ranking them based on their potential helpfulness. However, only a few reviews ever receive any helpfulness votes on online marketplaces. Sorting all reviews based on the few existing votes can cause helpful reviews to go unnoticed because of the limited attention span of readers. The problem of review helpfulness prediction is even more important for higher review volumes, and newly written reviews or launched products. In this work we compare the use of RoBERTa and XLM-R language models to predict the helpfulness of online product reviews. The contributions of our work in relation to literature include extensively investigating the efficacy of state-of-the-art language models---both monolingual and multilingual---against a robust baseline, taking ranking metrics into account when assessing these approaches, and assessing multilingual models for the first time. We employ the Amazon review dataset for our experiments. According to our study on several product categories, multilingual and monolingual pre-trained language models outperform the baseline that utilizes random forest with handcrafted features as much as 23\% in RMSE. Pre-trained language models reduce the need for complex text feature engineering. However, our results suggest that pre-trained multilingual models may not be used for fine-tuning only one language.

We assess the performance of language models with and without additional features. Our results show that including additional features like product rating by the reviewer can further help the predictive methods.
\end{abstract}
\raggedbottom
\section{Introduction} \label{sec:Intro}
Online reviews are a good source of information for prospective customers, helping them decide about purchasing a product or service. Reviews can also be helpful for businesses as a source for their products/services assessment by customers and to identify their strengths and weaknesses for better planning. As more customers adopt online shopping, the number of online customer reviews grows daily. Customers and businesses can become overwhelmed by the number of reviews for products/services. Hence, there is a need to identify the most helpful reviews. Specifically, sorting reviews based on their helpfulness can be very useful for both customers and businesses.

Some online marketplaces like Amazon have a voting feature allowing readers to specify whether a review is helpful. However, only a small fraction of reviews get any helpfulness votes since each customer usually reads only a few reviews, and many do not vote for reviews’ helpfulness. Simply sorting reviews based on their existing votes can exacerbate the problem. Most customers will only look at the few reviews with existing votes, resulting in potentially helpful reviews being neglected or overlooked with no votes~\cite{wan2015matthew,singh2017predicting}. This issue can be more prominent for products/services with a high review volume, more recent reviews, or for newly launched products.

In literature, the helpfulness of reviews is defined based on the helpfulness votes. In most of the works, reviews are defined as most helpful by the ratio of people that voted for the specific review as helpful over the total number of votes (helpful or not helpful)~\cite{yang2015semantic,singh2017predicting,mukherjee2017exploring,chen2018cross}. However, some datasets may lack the total number of votes for each review. In such cases, researchers have considered helpfulness as the number of votes a review has received for being helpful, possibly normalized by the elapsed time between review publication and dataset collection~\cite{tsai2020improving,bilal2022effectiveness}.

In this research, we aim to investigate the usefulness of models based on pre-trained language models to predict the helpfulness of an online review, even before the review has any helpfulness vote. An effective predictive method provides a fair chance of visibility to all reviews based on their potential helpfulness. In this research, we intend to compare methods based on fine-tuning state-of-the-art (SOTA) language models, specifically RoBERTa~\cite{liu2019roberta} and a multilingual pre-trained language model, XLM-R~\cite{conneau2020unsupervised}, and some more classical methods combined with handcrafted feature constructors.
We conducted experiments with English data collected from reviews submitted for Amazon products.

Here, we aim to investigate how the method based on the pre-trained multilingual model compares against the one based on the monolingual language model on reviews written in English, a very common language. Given that a method based on multilingual language models is more generally applicable, it can provide additional benefits in case reviews written in multiple languages. Moreover, these models can be fine-tuned on a larger dataset and generalized better to low-resource languages.

Our work investigates aspects not previously studied in depth for product review helpfulness prediction. Specifically, we examine the use of SOTA language models, both monolingual and multilingual models, for this task and consider ranking metrics in evaluating the methods. Thus, this work can guide us toward selecting and developing new models for sorting the reviews on online stores based on their predicted helpfulness.

We would like to emphasize that review helpfulness prediction is a different problem from the typical task of sentiment analysis. Merely comprehending that a review has neutral, positive, or negative sentiment is not the solution for finding and illustrating the most helpful reviews to the customers, as helpful reviews can be located among reviews with all types of sentiments. As the final product, we desire a method of sorting reviews that is more advantageous for the customers by better predicting the most helpful ones. A reliable review helpfulness predictor can assist customers in making more informed purchasing decisions while saving time. Also, businesses can grasp the essential customer needs and opinions more easily to be able to address them by adjusting their product or service according to society's needs.

This paper is organized as follows. We first cover the related work in Sect.~\ref{sec:literature}. In Sect.~\ref{sec:contribution}, we highlight our main contributions and the importance of the research. Sect.~\ref{sec:methods} presents our proposed methodology – language model-based methods. Empirical evaluations, including details about the dataset, the baseline models, evaluation metrics, and the results, are presented in Sect.~\ref{sec:empirical}. Sect.~\ref{sec:discussion} discusses the overall picture that emerges from the findings. Sect.~\ref{sec:conclusion} concludes the paper and gives insights for future work.

\section{Related Work}
\label{sec:literature}

Researchers have used various algorithms and features on different datasets like Amazon, TripAdvisor, and Yelp~\cite{yang2015semantic,singh2017predicting,sun2019helpfulness,bilal2022effectiveness}. We divide the previous works into three categories: regression, classification, and ranking, based on their approach to dealing with helpfulness prediction.

\subsection{Regression}

The works with a regression approach try to predict a number on a continuous scale (typically in [0,1]), indicating a review's helpfulness.
Researchers have mainly used support vector regression for the regression approaches~\cite{diaz2018modeling}. We can also see linear regression~\cite{lu2010exploiting}, probabilistic matrix factorization~\cite{tang2013context}, and extended tensor factorization models~\cite{moghaddam2012etf}, which have incorporated contextual information for authors or readers and have achieved better results than simpler regression models. However, depending on non-textual features may be limiting for many applications where that information is not collected.

The work presented in~\cite{yang2015semantic} isolates review helpfulness prediction from non-text features. They represent reviews in two semantic dimensions: linguistic inquiry and word count (LIWC)~\cite{pennebaker2007linguistic,boyd2022development} and INQUIRER\footnote{INQUIRER dictionary is introduced originally in~\cite{stone1962general}.}. LIWC considers semantic features to map text to numerical values representing emotions, writing styles, and other psychological and linguistic dimensions. Their findings reveal that the two semantic feature sets may more precisely predict helpfulness scores and significantly boost performance compared to utilizing features proposed prior to their work like geneva affect label coder (GALC)~\cite{martin2014prediction} and structural features~\cite{xiong2011automatically}. Their work motivates us to use LIWC features for the baseline.

Reference~\cite{mukherjee2017exploring} utilizes Hidden Markov Model – Latent Dirichlet Allocation to jointly model reviewer expertise as latent time-evolving variables and product aspects as latent topics. The authors give their attention to the semantics of reviews and explain how experts employ salient terms from latent word clusters to describe key aspects of the reviewed item. 
Their method outperforms the selected baselines from previous works~\cite{o2009learning,lu2010exploiting} for helpfulness prediction. One drawback of their work is that the method heavily depends on having access to multiple reviews from the same users over time, which may only sometimes be possible due to data collection and availability limitations.

The work presented in~\cite{chen2018cross} uses word and character representation to propose a convolutional neural network model to address cases with limited data and out-of-vocabulary (OOV) problem. Additionally, in order to predict the cross-domain review helpfulness, they jointly model product domains to transfer knowledge from a domain with more data to the one with less data. They use Amazon product review dataset and achieve better results in terms of correlation coefficient and also cross-domain robustness than the competing methods including~\cite{yang2015semantic} and~\cite{liu2017using}.
It is noteworthy that with the advent of more advanced language models, OOV may be less of an issue.

\subsection{Classification}

Methods with a classification approach attempt to classify the reviews as helpful or not helpful.
Most researchers have used support vector machines (SVMs) for the classification approach~\cite{diaz2018modeling}.
Reference~\cite{krishnamoorthy2015linguistic} proposes a model for helpfulness prediction extracting linguistic features, and making a feature value by accumulating them. The author concludes that these linguistic features help predict the helpfulness of reviews for experience products like books. Nevertheless, using an element of review metadata as a review rating can add noise due to higher misalignment with helpfulness. Therefore, they suggest designing a review ranking mechanism depending on the product type. However, some other features, like semantic features of review, are not included in their predictive model. Some newer works have dealt with classification using neural networks~\cite{malik2017helpfulness}.

Reference~\cite{sun2019helpfulness} investigates the impact of different review informativeness measurements on review's helpfulness. They use thresholded regression and mention that the classification threshold for search products such as laptops is more significant than experience products such as running shoes because search product attributes can be compared more manageably and objectively. They conclude that review informativeness plays a crucial role in the helpfulness of reviews. Their research does not include features related to the sentiment, which has been suggested by other works to play an essential role in determining review helpfulness. Our baseline includes features related to sentiment.

While automatic feature extraction for natural language processing (NLP) can be a significant step in review helpfulness analysis, very few recent works have used BERT~\cite{kenton2019bert} as one of the best-performing methods for this task~\cite{bilal2022effectiveness}.
Reference~\cite{bilal2022effectiveness} uses the Yelp shopping review dataset to compare the performance of a BERT-based classifier with the bag-of-words approaches combined with SVM or K-nearest neighbors (KNN). They consider reviews with 0 helpful votes unhelpful and reviews with 4 or more helpful votes helpful. One may argue that this labeling of helpful versus unhelpful is very subjective. Nevertheless, the authors conclude that in distinguishing helpful and unhelpful reviews, BERT-based classifiers outdo bag-of-words methods.
One downside of this work is the weak baseline based on simplistic bag-of-words representations. Moreover, it leaves evaluating the effectiveness of more recent language models and analysis on other datasets like product reviews to future work. To the best of our knowledge, no prior work has examined the utility of multilingual language models for review helpfulness prediction.

\subsection{Ranking}

Very few works have considered the problem directly as a ranking problem. Reference~\cite{tsur2009revrank} proposes an unsupervised method to rank reviews according to their distance to a ``virtual core review" which is identified as a lexicon of dominant terms in the reviews. But their approach requires an external reference corpus and is specifically tailored to book reviews only. Although ranking metrics are very relevant for evaluating review helpfulness prediction, works like~\cite{mukherjee2017exploring} and~\cite{hong2012reviews} that report a ranking-related metric along the usual regression or classification metrics are rare. To the best of our knowledge, no work using modern language models for helpfulness prediction has investigated this aspect and ranking metrics. 

\section{Research Contribution}
\label{sec:contribution}

Most of the previous works on predicting the helpfulness of the reviews use machine learning techniques such as linear regression, support vector machines, and random forest combined with features constructed from the reviews’ texts and possibly utilize additional information like star ratings. Text-based features can include word-based features. For example, TF-IDF, review length, presence of keywords or words from pre-defined lists of words relevant to sentiments/subjectiveness, and more complex semantic features like LIWC. Moreover, very few works have been done on product review helpfulness prediction using deep learning or cutting-edge language modeling approaches. However, this line of research has compared these new approaches with the classical ones using only more superficial text-based features like TF-IDF and not the more complex handcrafted ones that we employ in our baseline.

Therefore, in this research, \textit{we want to find out how the performance of a SOTA multilingual pre-trained language model (MLLM) like XLM-R compares with a SOTA pre-trained monolingual model (LM) such as RoBERTa and a classical approach, namely random forest, with handcrafted features in predicting helpfulness of product reviews}. In addition, the comparisons in the works mentioned above have been primarily based on using only the reviews' text. Thus, we want to investigate to what extent including additional features like star rating and review length helps the predictive models that are based on pre-trained language models for the task of review helpfulness prediction. Moreover, as some of the works discussed in the previous section have suggested that methods' results can vary on different product categories, especially between experience and search products. We also investigate how the relative performance of the methods we compare changes across different product categories.

\section{Proposed Methodology}
\label{sec:methods}

\subsection{Language Model-Based Methods}

This work uses SOTA pre-trained language models to build predictive models for review helpfulness prediction. We leverage these models to form (trainable) representations of the text of the reviews, which will then be fed to the predictor module.

Since the advent of BERT~\cite{kenton2019bert}, most of the new developments in language modeling have been primarily composed of transformers~\cite{vaswani2017attention}. Transformers are based on the multi-head (self) attention mechanism, a different paradigm from recurrent or convolutional networks, based on efficient feed-forward networks for translation and a positional encoder for predicting the word-order.
BERT is a bidirectional transformer pre-trained using a combination of next-sentence prediction and masked language modeling on a large unlabeled corpus of Toronto Book and Wikipedia. BERT learns representations of text by conditioning on both left and right context. In~\cite{kenton2019bert}, it has been shown that fine-tuning BERT with just one output layer results in a promising performance in several benchmark natural language understanding (NLU) tasks. BERT is often credited with marking a new era of NLP and making pre-training of language models a standard. The language models utilized in this work are based on transformers and improvements over BERT.

There have been several works that have been improvements of BERT. One that we consider in our work is RoBERTa~\cite{liu2019roberta}. RoBERTa discards the next sentence pre-training objective of BERT, trains the model on a larger corpus, and modifies important hyperparameters like using larger batch sizes and learning rates. It is shown in~\cite{liu2019roberta} that RoBERTa achieves SOTA results in the NLU benchmarks compared with all works prior to it.

Multilingual models based on transformers have also been developed in recent years. One of the SOTA models is XLM-R~\cite{conneau2020unsupervised}, a large multilingual masked language model based on RoBERTa. XLM-R was pre-trained on a large dataset (2.5 TB) from CommonCrawl~\cite{wenzek-etal-2020-ccnet} and one hundred languages. XLM-R has been shown to outperform other multilingual models on various cross-lingual benchmarks and even surpass the per-language performance of BERT monolingual baselines~\cite{conneau2020unsupervised}. Thus, in this work, we put XLM-R to test for the task of review helpfulness prediction.

\begin{figure}
    \centering
    \includegraphics[width=0.6\textwidth]{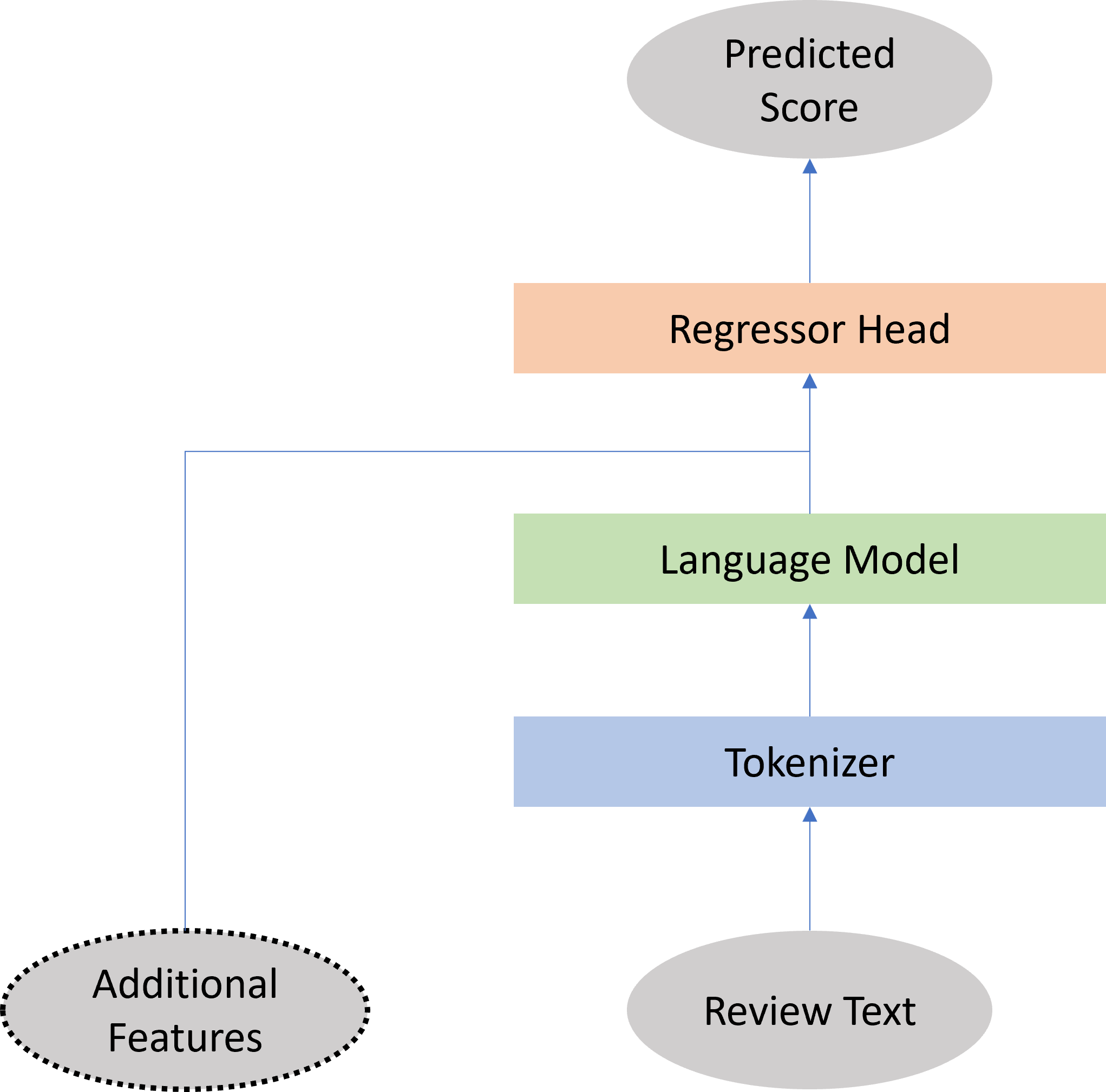}
    \caption{Overall model structure}
    \label{fig:model_architecture}
\end{figure}

We utilize a pre-trained language model to extract a representation from the review text and fine-tune the language and regression model with our review helpfulness data. The overall model structure is shown in Fig.~\ref{fig:model_architecture}.

We choose a tokenizer corresponding to the specific language model. We take the embedding of the start-of-sequence token (also known as classification token – CLS) as the representation of review text. One can also take the concatenation of embeddings of the start of the sequence token from the last couple of layers instead or do an average pooling over the embeddings of all tokens in the last layer. In our limited ablation tests, we saw no improvement from these strategies over using only the start of sequence token (CLS) embedding during fine-tuning, as we see in Table~\ref{tab:cls vs avg pooling}. Hence, from computational perspective, we choose to use CLS embedding, a 768-dimensional vector.

\begin{table}[ht!]
    \caption{Results of ablation tests, using only the start of sequence token (CLS) embedding during fine-tuning compared with average pooling over the embeddings of all tokens in the last layer. Results are based on the average of metrics for 3 different random splits of data for all 6 metrics on the cellphone category. Numbers in parentheses denote standard deviation (MAE: mean absolute error, RMSE: root mean squared error, PCC in [-1,1]: Pearson correlation coefficient, SPC in [-1,1]: Spearman's rho, KC in [-1,1]: Kendall's tau, NDCG in [0,1]: normalized discounted cumulative gain).}
    \centering
    \resizebox{1\textwidth}{!}{
    \begin{tabular}{ccccccccc}
        \hline
        Category & Model & Embedding Method & MAE & RMSE & PCC & SPC & KC & NDCG\\
        % \hhline{|=|=|=|=|=|=|=|=|=|}
        \hline
        \hline 
        Cellphone & RoBERTa & CLS & \textbf{0.1380} & 0.1990 & \textbf{0.6092} & 0.5215 & 0.3690 & \textbf{0.9787}\\
         & & & (0.0053) & (0.0033) & (0.0026) & (0.0164) & (0.0122) & (0.0009) 
         \\
        \hline
        Cellphone & RoBERTa & Average Pooling & 0.1396 & \textbf{0.1985} & 0.6062 & \textbf{0.5250} & \textbf{0.3721} & 0.9777\\
         & & & (0.0039) & (0.0041) & (0.0067) & (0.0158) & (0.0117) & (0.0007)
        \\
        % \hhline{|=|=|=|=|=|=|=|=|=|}
        \hline
        \hline
        Cellphone & XLM-R & CLS & 0.1429 & \textbf{0.1998} & \textbf{0.5997} & \textbf{0.5092} & \textbf{0.3595} & \textbf{0.9784}\\
         & & & (0.0054) & (0.0014) & (0.0072) & (0.0164) & (0.0112) & (0.0012)
        \\
        \hline
        Cellphone & XLM-R & Average Pooling & 0.1429 & 0.2009 & 0.5916 & 0.5080 & 0.3585 & 0.9780\\
         & & & (0.0022) & (0.0030) & (0.0067) & (0.0152) & (0.0103) & (0.0007)
        \\
        \hline
    \end{tabular}}
    \label{tab:cls vs avg pooling}
\end{table}

We use two fully-connected layers for the regressor head, with the first one having a nonlinear ReLU activation function when utilizing additional features and a single layer otherwise. We do not want to restrict the relationship between additional features and output to a simple linear one. When using features in addition to the review text, they are concatenated with the text embedding and fed to the regressor head.

In this work, we investigate whether models based on language models can benefit from additional features like star rating and review length (in terms of number for words) for review helpfulness prediction. We normalize these two additional features before concatenation with CLS embedding by centering and scaling them based on statistics computed from training data.

\begin{figure}
    \centering
    \includegraphics[width=0.9\textwidth]{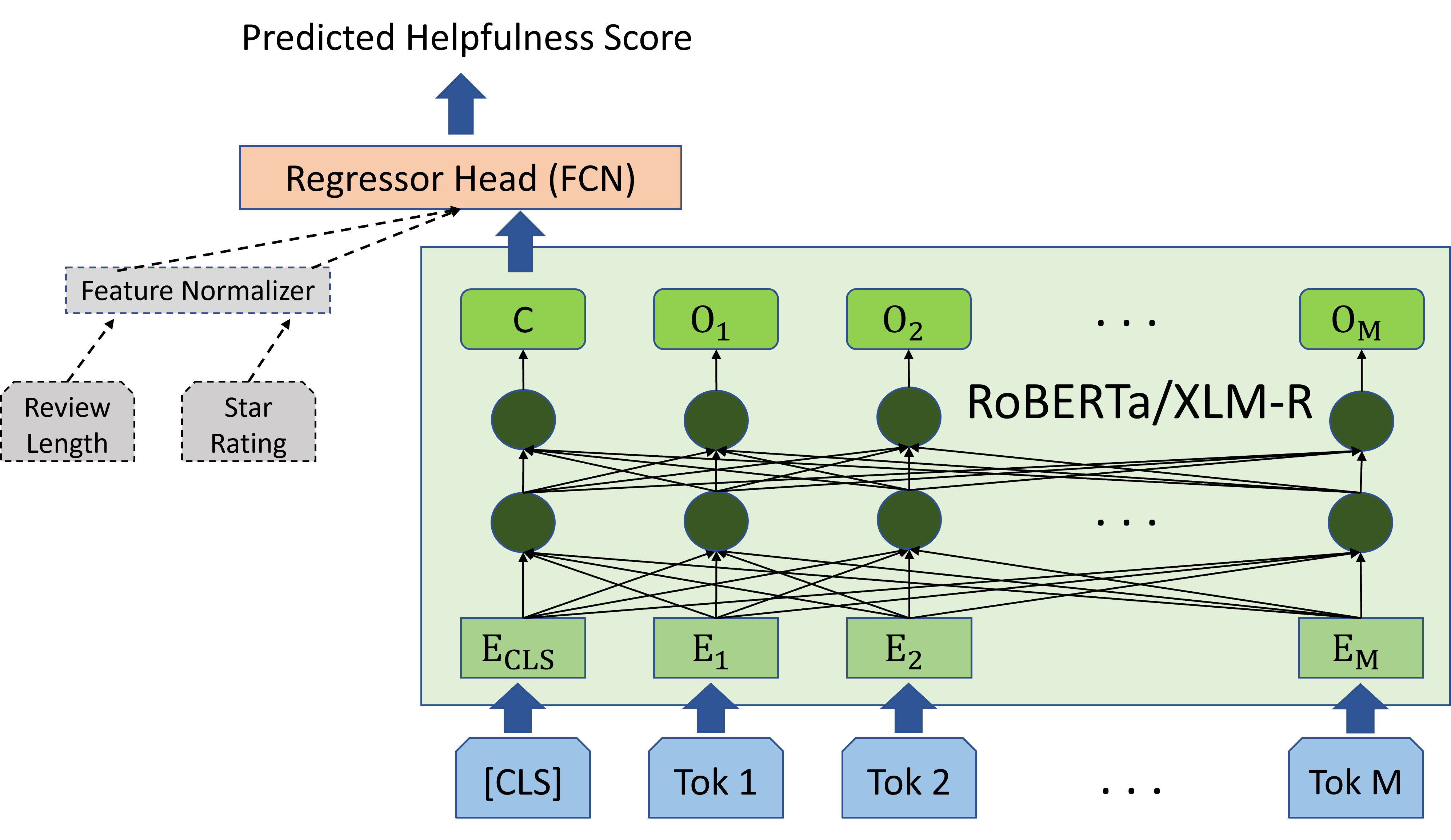}
    \caption{Architecture of language model-based methods. Tok stands for token, E and O denote the initial and output embeddings, the regressor head is an FCN (fully-connected network), and M represents the maximum token sequence length set to 300 in this work. When using additional features, the embedding of the start of the sequence is concatenated with additional features and then fed to the regressor head.}
    \label{fig:model_architecture_detail}
\end{figure}

To select the maximum length of token sequences, we computed the 75\% quantile of tokenized reviews' length for each category and computed their average, which resulted in a number close to 300. Motivated by this, and to make the models and their training less computationally expensive, as the computational complexity of attention-based models is quadratic in sequence length, we set the maximum token sequence length to 300 instead of the maximum possible value of 512. A more detailed view of the predictive models based on pre-trained language models is shown in Fig.~\ref{fig:model_architecture_detail}.

As the problem is of regression type, the model is trained with mean-squared-error (MSE) loss. Following~\cite{liu2019roberta,conneau2020unsupervised,kenton2019bert}, we employ Adam optimizer~\cite{kingma2015adam} with $\beta_1=0.9$,  $\beta_2=0.999$, learning rate warmup over the steps of the first epoch to peak learning rate value and linear decay afterward, and batch size of 16. We did some ablation tests to select the peak learning rate value from [1e-4, 1e-5] and observed that 1e-4 shows slightly better performance. For each training run, 12.5\% of the products in training data (equivalent to 10\% of all products) are set aside for validation set. We train the model for 5 epochs, monitor the model's performance on the validation set during training, and restore the model state from the epoch with the best performance on the validation set.

Facebook has originally developed both RoBERTa and XLM-R, and their official implementation in PyTorch and pre-trained versions are available at [\footnote{\url{https://github.com/pytorch/fairseq/tree/master/examples/roberta}}] and [\footnote{\url{https://github.com/pytorch/fairseq/tree/master/examples/xlmr}}]. Both of these models are also implemented in the transformers class of HuggingFace\footnote{\url{https://huggingface.co/}} and are available for both PyTorch and TensorFlow. We use the transformers class with PyTorch in this study. The rest of the hyperparameters not discussed in this section are set to their suggested default values in HuggingFace. Experiments are run on a cluster node with Intel Xeon 6346 3.10 GHz processor and an NVIDIA A40 Accelerator.

\section{Empirical Evaluations}
\label{sec:empirical}

\subsection{Dataset Description and Processing}

As noted in Sect.~\ref{sec:Intro}, we work on the Amazon review dataset, which includes reviews (ratings, text, helpfulness votes, product ID, reviewer ID), and product metadata (category information, price, brand) provided by McAuley~\cite{he2016ups} and downloadable at:  [\footnote{\url{http://jmcauley.ucsd.edu/data/amazon/links.html}}]. One of the advantages of this dataset is that not only it contains the number of helpful votes that a review gets, but it also includes the total number of votes. This allows us to look at our prediction task with less subjectivity and from a regression perspective without needing to set subjective thresholds or scaling.

We use four of the original dataset columns, which are:
(i) Review Text: contains the text of the review.
(ii) Overall: specifies the product's rating by the reviewer.
(iii) Asin: indicating the product ID.
(iv) Helpful: includes the number of people who find the review helpful and the total number of votes. We construct the helpfulness ratio column out of this last original column.

We choose four product categories from the dataset: beauty products, movies, cellphones, and electronics, to have a mix of categories in terms of experience vs. search and data size. The pre-processing steps we take are as follows.
First, we remove the rows (data points) that do not have any value for review text, overall, or helpful column. In order to have a more stable and less biased helpfulness score (ratio), we follow the common best practices~\cite{park2018predicting,chen2018cross,yang2015semantic} and filter the reviews to keep the ones with more than 10 votes from our dataset.

Although we can expect to only keep meaningful reviews by dropping the ones with less than 10 votes, we further check the remaining data for reviews without any alphabetical words. After filtering, we have 883,412 reviews from the four selected categories. In Table~\ref{tab:data-summary}, we can see the number of reviews for each of the four categories. Helpfulness scores are between 0 and 1 (inclusive).

\begin{table}[h]
\caption{Properties of the different categories of the dataset}
    \centering
    \resizebox{1\textwidth}{!}{
    \begin{tabular}{ccccc}
        \hline
        Category &  Electronics  &  Beauty Products  &  Cellphone  &  Movies \\
        \hline
        \hline
        No. Reviews Before Filtering & 7,824,482 & 2,023,070 & 3,447,249 & 4,607,047\\
        \hline
        No. Reviews After Filtering & 359,881 & 59,359 & 53,854 & 410,318
        \\
        \hline

    \end{tabular}}
    \label{tab:data-summary}
\end{table}

\subsection{Baseline}

One advantage of pre-trained language models is that there is usually no need to find additional handcrafted features from text for downstream tasks. However, as our baseline, we include a more classical machine learning-based way of predicting helpfulness. Specifically, instead of a language model, we utilize handcrafted features from review texts that have been shown to work well among the different features tested in the literature. We use LIWC, a software program to analyze text and convert it into a structured format with numerical scores based on its internal dictionary about different psychological and linguistic features~\cite{pennebaker2007linguistic}. In this work, we employ LIWC-22~\cite{boyd2022development}, the fifth upgraded version of the original program with more extensive dictionaries.

The core of LIWC is its dictionary. Its inner lexicon comprises around 12,000 words, word stems, idioms, and several emoticons. Each item in the dictionary is a component of one or more sub-dictionaries or categories, mostly organized hierarchically and intended to evaluate different psychological variables. LIWC assigns zero or one to each word in the review text in different language dimensions. After summing up the values of all words for each dimension, we will have a histogram of language dimensions for each review.

We utilize LIWC and process every review text to get its representation in LIWC dimensions. Overall, LIWC features act as a baseline for review text representation. We use a more classical machine learning method, Random Forest Regressor, and its implementation in scikit-learn\footnote{\url{https://scikit-learn.org/}} for this baseline. This baseline follows the construction used in~\cite{chen2018cross,park2018predicting,yang2015semantic}. For our baseline, we utilize all the available dimensions in LIWC. For hyperparameter tuning, we consider the following hyperparameters and possible values for them: \{\texttt{n\_estimator}: [200, 400], \texttt{max\_features}: ['auto', 'sqrt'], \texttt{max\_depth}: [10, None], \texttt{min\_samples\_leaf}: [10, 50]\}. \texttt{n\_estimator} determines the number of trees, \texttt{max\_features} specifies the number of features to consider when searching for the best split at each node, where 'auto' means using all features. \texttt{max\_depth} is the maximum possible tree depth with 'None' imposing no constraint, and \texttt{min\_samples\_leaf} is the required minimum number of samples at each leaf node.

\subsection{Evaluation Method}

We first split the dataset into training/validation and test sets to evaluate the different methods. The splitting strategy has been either entirely random or considered products in previous works. Following~\cite{mukherjee2017exploring,fan2019product}, we split the dataset of each product category to 80\% of products for training and validation and 20\% of products held out for testing. We split the data by product so that the model does not see any review for the products in the test set during training. This strategy of splitting the data is closer to an actual application of a review helpfulness prediction model compared with a completely random splitting. It is noteworthy that the splitting by product setup may be more challenging for the model than an utterly random splitting setup, since in the latter the model can see labeled reviews for the products in the test set during training. For most of the comparison results, we do three different splits of training/validation and test data to compute statistics of performance metrics.
We use standard evaluation metrics for regression which are root mean squared error (RMSE), mean absolute error (MAE), and Pearson correlation coefficient (PCC).

As mentioned earlier in Sect.~\ref{sec:Intro}, the ground truth of the helpfulness score is defined by the ratio of people that have voted for the specific review as helpful over the total number of votes (helpful or not helpful)~\cite{yang2015semantic,mukherjee2017exploring,singh2017predicting,chen2018cross}. That is, $y=\frac{n_{helpful}}{n_{total}}$, with $n_{helpful}$ and $n_{total}$ representing the number of helpful votes and the number of total votes, respectively. PCC is a value between -1 and 1, with higher values indicating the better performance of the model predictions. In addition to the usual regression metrics, we consider ranking metrics for evaluation purposes. Although a marketplace ultimately needs to rank the reviews based on their potential helpfulness, only a few works have reported results related to ranking metrics. Here, we consider ranking metrics such as Spearman's rho (SPC), Kendall's tau (KC), and normalized discounted cumulative gain (NDCG) to compare rankings of reviews based on the prediction of the model and the ground truth.

SPC is the non-parametric version of PCC and assesses the monotonic association between two variables and returns a value between -1 and 1, with higher values suggesting a stronger association. SPC can be calculated as the PCC of ranks of $y$ and $\hat{y}$. KC is another measure for correspondence between two rankings, measuring the ordinal association between variables, and is a value between -1 and 1, with values closer to 1 indicating a stronger association. KC is calculated based on the number of pairs in the same order (concordant) and the number of pairs in a different order (discordant) in the two rankings. While PCC measures the linear relationship between prediction and ground truth, SPC and KC measure more general monotone associations.

\begin{figure}
    \centering
    \includegraphics[width=0.94\textwidth]
    {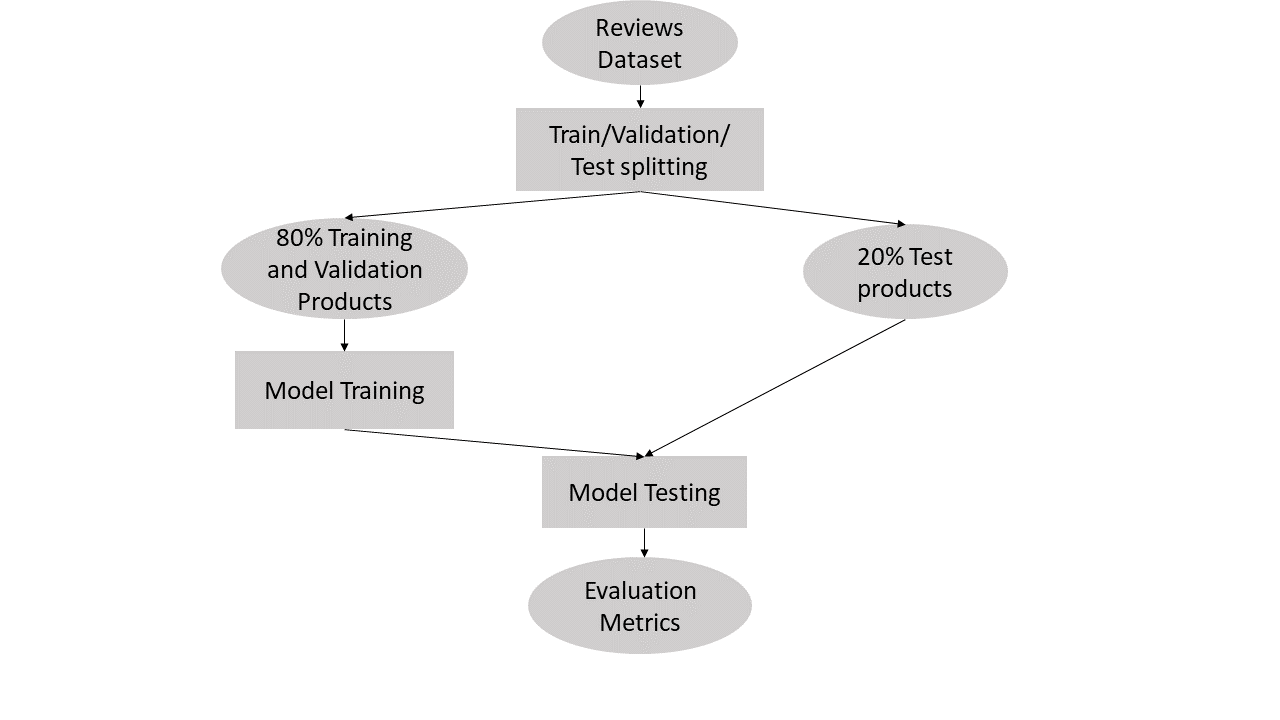}
    \caption{Experiment process}
    \label{fig:flowchart_eval}
\end{figure}

NDCG is a popular metric for comparing ranking systems in information retrieval. NDCG for a rank position is the DCG through that rank position normalized by the ideal maximum possible DCG through that rank position. DCG is the sum of true relevance scores ranked in the order induced by predicted relevance scores, with a logarithmic discount that increases with the rank position. Thus, NDCG weights the top-rank positions more than the lower ones and is a value between 0 and 1, with 1 being the best score. In our case, helpfulness scores are, in fact, the relevance scores. As noted in the literature, the users usually only read a limited number of reviews, so NDCG@k, which only considers the top k for computing the metric, may be the most suitable for our application. Hence, in this work, we compute NDCG@10 for each product in the test set and report the average across products. The flowchart of the overall evaluation process is shown in Fig.~\ref{fig:flowchart_eval}.

\subsection{Results}

First we conduct an ablation study to find the optimal setting for fine-tuning the language models and present the results below. Then, we provide the results of the models based on pre-trained language models and the baseline for different categories of the products.

\subsubsection{Ablation Study on Fine-Tuning:}

The architectures of RoBERTa and XLM-R that we use in this work consist of 12 attention layers. Motivated by prior works such as~\cite{lee2019would,eberhard2021effects} that suggest freezing a subset of layers may be beneficial depending on the task and dataset, we first perform an ablation study to determine how many layers of the pre-trained language model need to be fine-tuned to achieve the best performance that is the best trade-off between performance and computational complexity. Fine-tuning fewer layers can speed up training while potentially reducing the chances of overfitting.

Since the pre-training procedure of RoBERTa and XLM-R does not involve the next sentence prediction, there is no justification for freezing all layers of the language model and using the CLS embedding as the text representation without any fine-tuning of the model. Thus, we need to fine-tune a subset of the layers at least. In deep learning models, specifically in NLP and computer vision, a common belief is that the initial layers encode more universal features. In contrast, the top/last layers encode more task-specific ones~\cite{kovaleva2019revealing,clark2019does,zeiler2014visualizin}. Hence, the freezing of the layers is always done in a bottom (initial layers) to top (last layers) fashion.
We train our RoBERTa-based and XLM-R-based models by fine-tuning the top (last) 2, 4, 6, and all of the layers of the language model. The cutoff of 6 layers for exploration of fine-tuning the different number of layers is inspired by~\cite{lee2019would}, where they suggest that the first half of the language model layers can be confidently kept frozen considering the accuracy and computational complexity. It is worth mentioning that in all cases, the regressor head is trained during the fine-tuning procedure.

Table~\ref{tab:number_of_layers} shows the results of training models based on pre-trained RoBERTa and XLM-R, with the different number of layers being fine-tuned for cellphone and movie categories. We choose these two categories for this test to have categories with different sample sizes and product types.
The results are based on 3 different splits of train/validation and test data, and the reported numbers are the metrics computed on test data and averaged across the 3 runs.

\FloatBarrier
\begin{table}[h]

\caption{Results with changing the number of fine-tuned layers in the models based on pre-trained language models. Results are based on the average of metrics for 3 different random splits of data for all 6 metrics on cellphone and movie categories.}

    \centering
    \resizebox{1\textwidth}{!}{
    \begin{tabular}{ccccccccc}
        \hline
        Category & Model & No. Lay. & MAE$~\downarrow$ & RMSE$~\downarrow$ & PCC$~\uparrow$ & SPC$~\uparrow$ & KC$~\uparrow$ & NDCG$~\uparrow$\\
        % \hhline{|=|=|=|=|=|=|=|=|=|}
        \hline
        \hline 
        Movies & RoBERTa & 2 & \textbf{0.1475} & \textbf{0.2041} & \textbf{0.7742} & \textbf{0.7717} & \textbf{0.5717} & \textbf{0.9737}
         \\
        \hline
        Movies & RoBERTa & 4 & 0.1512 & 0.2061 & 0.7682 & 0.7656 & 0.5650 & 0.9733
         \\
        \hline
        Movies & RoBERTa & 6 & 0.1567 & 0.2136 & 0.7566 & 0.7541 & 0.5538 & 0.9723
         \\
        \hline
        Movies & RoBERTa & all & 0.1546 & 0.2142 & 0.7538 & 0.7516 & 0.5523 & 0.9720
        \\
        \hline
        Movies & XLM-R & 2 & \textbf{0.1564} & \textbf{0.2108} & \textbf{0.7545} & \textbf{0.7558} & \textbf{0.5563} & \textbf{0.9719}
        \\
        \hline
        Movies & XLM-R & 4 & 0.1615 & 0.2142 & 0.7446 & 0.7408 & 0.5408 & 0.9715
        \\
        \hline
        Movies & XLM-R & 6 & 0.1671 & 0.2240 & 0.7207 & 0.7176 & 0.5191 & 0.9697
        \\
        \hline
        Movies & XLM-R & all & 0.2230 & 0.2706 & 0.5633 & 0.5669 & 0.4024 & 0.9556
        \\
        \hline
        \hline
        Cellphone & RoBERTa & 2 & 0.1380 & 0.1990 & 0.6092 & 0.5215 & 0.3690 & 0.9787
         \\
        \hline
        Cellphone & RoBERTa & 4 & 0.1376 & \textbf{0.1967} & \textbf{0.6132} & \textbf{0.5240} & \textbf{0.3710} & 0.9785
        \\
        \hline
        Cellphone & RoBERTa & 6 & 0.1389 & 0.2010 & 0.5951 & 0.4977 & 0.3507 & 0.9787
        \\
        \hline
        Cellphone & RoBERTa & all & \textbf{0.1355} & 0.1988 & 0.6044 & 0.5124 & 0.3620 & \textbf{0.9791}
        \\
        \hline
        Cellphone & XLM-R & 2 & \textbf{0.1429} & \textbf{0.1998} & \textbf{0.5997} & \textbf{0.5092} & \textbf{0.3595} & 0.9784
        \\
        \hline
        Cellphone & XLM-R & 4 & 0.1460 & 0.2018 & 0.5915 & 0.4951 & 0.3486 & \textbf{0.9786}
        \\
        \hline
        Cellphone & XLM-R & 6 & 0.1458 & 0.2044 & 0.5691 & 0.4757 & 0.3338 & 0.9777
        \\
        \hline
        Cellphone & XLM-R & all & 0.1632 & 0.2159 & 0.5392 & 0.4612 & 0.3223 & 0.9777
        \\
        \hline

    \end{tabular}}
    \label{tab:number_of_layers}
\end{table}
\FloatBarrier

As we see in Table~\ref{tab:number_of_layers}, for the movie category, the average results show that with fine-tuning 2 layers, we have the best performance according to all of the 6 metrics for both RoBERTa-based and XLM-R-based models. Fine-tuning 4 layers is also better than all layers. This may indicate that the models can overfit the training data when fine-tuning more layers than needed, given the task and dataset size. For the cellphone category, fine-tuning 2 layers shows the best performance for the XLM-R-based model. For the RoBERTa-based model, although fine-tuning 4 layers shows the best performance overall, the results are very close to fine-tuning 2 layers. It is noteworthy that we also tested fine-tuning all layers with gradient clipping without any success in improving its result. Thus, according to the results of this comparison, and by considering computational complexity, and to have a fair and consistent comparison between the different models based on pre-trained RoBERTa and XLM-R, we choose to fine-tune 2 layers for both RoBERTa-based and XLM-R-based models as our default setting to investigate our research questions further.

\subsubsection{Results of Different Methods on All Product Categories:}

We do three random splitting of data to train, validate and test to have a better estimate of performances based on multiple independent runs and data splits. All the methods are tested on each split set; the metrics' average and standard deviation are computed across the independent runs. The average and standard deviation of all metrics for all methods are included in Table~\ref{tab:results_avg_std}.

\FloatBarrier
\begin{table}[ht!]
\caption{Results based on 3 runs on different random splits for all 6 metrics and product categories. RoBERTa Add and XLM-R Add are setups with additional features concatenated to review text representation. Numbers in parentheses denote standard deviation.}

    \centering
    \resizebox{1\textwidth}{!}{
    \begin{tabular}{cccccccc}
        \hline
        Category & Model & MAE$~\downarrow$ & RMSE$~\downarrow$ & PCC$~\uparrow$ & SPC$~\uparrow$ & KC$~\uparrow$ & NDCG$~\uparrow$
        \\
        % \hhline{|=|=|=|=|=|=|=|=|=|}
        \hline
        \hline
        Movies & RoBERTa & 0.1475 & 0.2041 & 0.7742 & 0.7717 & 0.5717 & 0.9737 \\
         & & (0.0009) & (0.0014) & (0.0036) & (0.0032) & (0.0028) & (0.0003)
        \\
        \hline
        Movies & RoBERTa Add & \textbf{0.1441} & \textbf{0.1983} & \textbf{0.7864} & \textbf{0.7810} & \textbf{0.5797} & \textbf{0.9752} \\
         & & (0.0007) & (0.0013) & (0.0025) & (0.0020) & (0.0016) & (0.0000)
        \\
        \hline
        Movies & XLM-R & 0.1564 & 0.2108 & 0.7545 & 0.7558 & 0.5563 & 0.9719 \\
         & & (0.0012) & (0.0036) & (0.0048) & (0.0010) & (0.0008) & (0.0003)
        \\
        \hline
        Movies & XLM-R Add & 0.1502 & 0.2017 & 0.7753 & 0.7676 & 0.5665 & 0.9737 \\
         & & (0.0011) & (0.0011) & (0.0026) & (0.0032) & (0.0034) & (0.0002)
        \\
        \hline
        Movies & RF & 0.2091 & 0.2563 & 0.5937 & 0.5873 & 0.4140 & 0.9603 \\
         & & (0.0017) & (0.0017) & (0.0017) & (0.0018) & (0.0012) & (0.0001)
        \\
        \hline
        \hline 
        Electronics & RoBERTa & \textbf{0.1338} & 0.2008 & 0.7104 & 0.6504 & 0.4732 & 0.9804
        \\
         & & (0.0007) & (0.0036) & (0.0071) & (0.0032) & (0.0030) & (0.0002) \\
        \hline
         Electronics & RoBERTa Add & 0.1339 & \textbf{0.1944} & \textbf{0.7220} & \textbf{0.6565} & \textbf{0.4774} & \textbf{0.9818} \\
          & & (0.0015) & (0.0014) & (0.0041) & (0.0000) & (0.0004) & (0.0003)
        \\
        \hline
         Electronics & XLM-R & 0.1392 & 0.2021 & 0.6975 & 0.6362 & 0.4610 & 0.9798 \\
          & & (0.0027) & (0.0013) & (0.0013) & (0.0024) & (0.0023) & (0.0003)
        \\
        \hline
         Electronics & XLM-R Add & 0.1345 & 0.1962 & 0.7157 & 0.6522 & 0.4737 & 0.9816 \\
          & & (0.0030) & (0.0028) & (0.0092) & (0.0093) & (0.0078) & (0.0005)
        \\
        \hline
         Electronics & RF & 0.1781 & 0.2366 & 0.5316 & 0.4768 & 0.3344 & 0.9737 \\
          & & (0.0009) & (0.0009) & (0.0049) & (0.0059) & (0.0042) & (0.0002)
        \\
        % \hhline{|=|=|=|=|=|=|=|=|=|}
        \hline
        \hline
        Beauty Prod & RoBERTa & 0.1313 & 0.1815 & 0.5719 & 0.4486 & 0.3165 & 0.9754  \\
        & & (0.0019) & (0.0028) & (0.0142) & (0.0008) & (0.0007) & (0.0005)
        \\
        \hline
        Beauty Prod & RoBERTa Add & 0.1314 & \textbf{0.1791} & \textbf{0.5877} & \textbf{0.4626} & \textbf{0.3267} & 0.9763 \\
         & & (0.0022) & (0.0043) & (0.0171) & (0.0101) & (0.0074) & (0.0007)
        \\
        \hline
        Beauty Prod & XLM-R & 0.1327 & 0.1841 & 0.5531 & 0.4372 & 0.3075 & 0.9750 \\
         & & (0.0030) & (0.0022) & (0.0132) & (0.0030) & (0.0023) & (0.0005)
        \\
        \hline
        Beauty Prod & XLM-R Add & \textbf{0.1305} & 0.1824 & 0.5679 & 0.4486 & 0.3155 & \textbf{0.9767} \\
         & & (0.0021) & (0.0023) & (0.0127) & (0.0100) & (0.0078) & (0.0006)
        \\
        \hline
        Beauty Prod & RF & 0.1477 & 0.1995 & 0.4134 & 0.3050 & 0.2110 & 0.9726  \\
        & & (0.0002) & (0.0009) & (0.0134) & (0.0024) & (0.0018) & (0.0005)
        \\
        % \hhline{|=|=|=|=|=|=|=|=|=|}
        \hline
        \hline
        Cellphone & RoBERTa & 0.1380 & 0.1990 & 0.6092 & 0.5215 & 0.3690 & 0.9787 \\
         & & (0.0053) & (0.0033) & (0.0026) & (0.0164) & (0.0122) & (0.0009) 
        \\
        \hline
        Cellphone & RoBERTa Add & \textbf{0.1344} & \textbf{0.1940} & \textbf{0.6330} & \textbf{0.5310} & \textbf{0.3759} & \textbf{0.9810} \\
         & & (0.0048) & (0.0037) & (0.0047) & (0.0059) & (0.0031) & (0.0006)
        \\
        \hline
        Cellphone & XLM-R & 0.1429 & 0.1998 & 0.5997 & 0.5092 & 0.3595 & 0.9784 \\
         & & (0.0054) & (0.0014) & (0.0072) & (0.0164) & (0.0112) & (0.0012)
        \\
        \hline
        Cellphone & XLM-R Add & 0.1361 & 0.1953 & 0.6189 & 0.5257 & 0.3719 & 0.9802 \\
         & & (0.0043) & (0.0030) & (0.0034) & (0.0146) & (0.0106) & (0.0008)
        \\
        \hline
        Cellphone & RF & 0.1633 & 0.2232 & 0.4375 & 0.3801 & 0.2628 & 0.9747 \\
         & & (0.0012) & (0.0026) & (0.0072) & (0.0114) & (0.0078) & (0.0012) 
        \\
        \hline

    \end{tabular}}
    \label{tab:results_avg_std}
\end{table}
\FloatBarrier

To specify the statistical significance of the differences between methods, we perform head-to-head \emph{t}-tests between different models based on their 3 runs' results to investigate further whether the differences are significant by considering \emph{p}-value of 0.05 and include the results in Table~\ref{tab:significance test}.

\FloatBarrier
\begin{table}[h!]
\caption{Statistical significance of head-to-head performance differences for all metrics and product categories. All methods are run and tested on 3 different splits of data. Y denotes that performance differences are significant, and N means not significant based on \emph{t}-test and considering \emph{p}-value of 0.05.}

\begin{center}
\resizebox{0.64\textwidth}{!}{
\begin{tabular}{ccccccc}
\hline
\multirow{2}{*}{Category}&\multicolumn{6}{c}{RoBERTa vs XLM-R}\\ \cline{2-7}
&MAE&RMSE&PCC&SPC&KC&NDCG\\ \hline
Movies & Y & N & Y & Y & Y & Y \\
Electronics & N & N & N & N & N & N \\
Beauty Prod & N & N & Y & Y & Y & N \\
Cellphone & N & N & N & N & N & N \\
\hline
\hline
\multirow{2}{*}{Category}&\multicolumn{6}{c}{RoBERTa vs RoBERTa Add}\\ \cline{2-7}
&MAE&RMSE&PCC&SPC&KC&NDCG\\ \hline
Movies & Y & Y & Y & Y & Y & Y \\
Electronics & N & N & Y & N & N & Y \\
Beauty Prod & N & N & Y & N & N & Y \\
Cellphone & N & Y & Y & N & N & Y \\
\hline
\hline
\multirow{2}{*}{Category}&\multicolumn{6}{c}{XLM-R vs XLM-R Add}\\ \cline{2-7}
&MAE&RMSE&PCC&SPC&KC&NDCG\\ \hline
Movies & N & N & Y & Y & Y & Y \\
Electronics & N & N & N & N & N & Y \\
Beauty Prod & N & Y & N & N & N & Y \\
Cellphone & N & N & N & N & N & N \\
\hline
\hline
\multirow{2}{*}{Category}&\multicolumn{6}{c}{RoBERTa Add vs XLM-R Add}\\ \cline{2-7}
&MAE&RMSE&PCC&SPC&KC&NDCG\\ \hline
Movies & Y & Y & Y & Y & N & Y \\
Electronics & N & N & N & N & N & N \\
Beauty Prod & N & N & N & Y & Y & N \\
Cellphone & N & N & N & N & N & N \\
\hline
\hline
\multirow{2}{*}{Category}&\multicolumn{6}{c}{RF vs XLM-R}\\ \cline{2-7}
&MAE&RMSE&PCC&SPC&KC&NDCG\\ \hline
Movies & Y & Y & Y & Y & Y & Y \\
Electronics & Y & Y & Y & Y & Y & Y \\
Beauty Prod & Y & Y & Y & Y & Y & Y \\
Cellphone & Y & Y & Y & Y & Y & Y \\
\hline

\end{tabular}
}
\label{tab:significance test}
\end{center}
\end{table}
\FloatBarrier

For the movies category, we see that the RoBERTa-based model outperforms the XLM-R-based one in all six metrics, and the differences are significant for nearly all metrics. We also observe that both models benefit from additional features with statistically significant differences. The RoBERTa-based model with additional features performs better than the XLM-R-based model with additional features. Comparing our baseline with XLM-R-based model, which has the worst performance among models based on pre-trained language models, we can see that the XLM-R-based model outperforms the baseline in all metrics, with significant differences. In short, we observe that RoBERTa-based model with additional features has the best performance, and the baseline has the worst performance in all metrics.

For the electronics category, the results indicate that the RoBERTa-based model's average metrics are better than the average metrics from the XLM-R-based model. However, metrics differences are not significant. Although additional features help models in this category, the differences are insignificant in most metrics. Comparing the model based on RoBERTa with additional features with the model based on XLM-R with additional features, we can see that the average metrics are very slightly better for RoBERTa. However, none of the differences is detected as significant. We again see that our baseline has the worst performance by far among all the models we tested. The best and worst performing models are the same as for the movies category.

For the beauty product category, we again observe that the RoBERTa-based model has better average metrics than XLM-R-based model. However, only the differences in correlation metrics are detected as significant. Comparing the RoBERTa-based model with and without additional features, it appears that additional features help the model, although only on two metrics are the differences identified as significant. The same holds for comparing XLM-R-based models with and without additional features. Moreover, four of the six average metrics for the RoBERTa-based model with additional features are better than the XLM-R-based model with additional features. At the same time, only two are detected as significant. We can again see that the baseline is significantly worse than all methods based on pre-trained language models. Overall, the best and worst performing methods for this category are the RoBERTa-based model with additional features and the baseline, respectively.

For the cellphone category, while the average metrics are better for the RoBERTa-based model compared to the XLM-R-based model, none of the differences are detected as significant. Models based on RoBERTa and XLM-R benefit slightly from additional features, especially for RoBERTa, where half the differences are recognized as significant. The RoBERTa-based model with additional features performs slightly better than the XLM-R-based model with additional features, albeit no metrics show as significant. All pre-trained language model-based methods significantly outperform the baseline. The best and worst performing methods are similar to other product categories.

\section{Discussion}
\label{sec:discussion}

This work provides an extensive comparison between models based on monolingual and multilingual language models and a robust classical baseline for review helpfulness prediction. Its results and analysis give insights into designing new methods for review helpfulness prediction and sorting reviews by employing SOTA language models. To the best of our knowledge, our work is the first to consider models based on pre-trained multilingual language models for review helpfulness prediction.

According to the results, we can confidently say that models based on pre-trained language models outperform the baseline that employs handcrafted features. Although BERT-based models have been shown to perform better than classical models using TF-IDF features~\cite{bilal2022effectiveness}, our results confirm that methods based on transformer-based language models can outperform more traditional machine learning methods even when using handcrafted features. Additionally, the results suggest that models based on RoBERTa perform slightly better than models based on XLM-R for the review helpfulness task under study. i.e., the performance orders among the results are on par across all categories. But only in one product category the difference is strongly statistically significant.
Moreover, while previous works have mentioned that pre-trained multilingual models can sometimes perform even better than monolingual models for a specific task and target language~\cite{doddapaneni2021primer,conneau2020unsupervised}, in this work, which compares for the first time these models for review helpfulness prediction, we do not observe any improvement from XLM-R compared with RoBERTa on reviews in English.

One advantage of the pre-trained language models is that they learn complex and informative text representations. Our results indeed show that they do not require additional out-of-text features to perform reasonably. Nevertheless, to explore whether additional features can still be beneficial, we add two additional features: the star rating that the person who wrote the review gave to the product, and word count, which is the total number of words in the review. The results from our experiments show that adding the additional features can improve the performance of both RoBERTa and XLM-R-based models in all the metrics. Although the degree of result improvement varies for different product categories from significant (e.g. movies) to non-significant (e.g. electronics), there is no case where additional features hurt the performance.

Moreover, we do not observe any meaningful change in the relative ordering of the models' performances across the different product categories. i.e., general conclusions can be drawn regardless of the product category. This is in line with previous works like~\cite{singh2017predicting} and somewhat disagrees with~\cite{krishnamoorthy2015linguistic}. We only see the significance level of differences changing across product categories. The model based on RoBERTa with additional features shows the best performance among all the investigated models, and our baseline, the random forest, has the worst performance.

\section{Conclusion and Future Work}
\label{sec:conclusion}

In this work, we extensively analyzed the utility of SOTA open-source pre-trained language models, both monolingual and multilingual, for review helpfulness prediction. We observe that these models outperform classical approaches, even when classical methods are strengthened by employing handcrafted features from subscription-based software programs. Our experiments indicate that when the fine-tuning and testing are on reviews from the same language, one may not see any improvement from pre-trained multilingual models over monolingual ones. Furthermore, our results suggest that despite the good performance of models based on pre-trained language models, they can still slightly benefit from including additional features like the review's star rating. Another observation is that the relative performance of different methods is consistent across different product categories.

Identifying the most helpful reviews, even before they have any helpfulness vote, and ranking reviews accordingly for customers is advantageous. It helps customers avoid going through many reviews to gain helpful information and perspectives. This can save customers' time, increase their satisfaction, and help them make informed decisions, resulting in more sales and/or lower return rates, eventually benefiting the marketplace and businesses. Businesses can also easily spot their products' strengths and weaknesses from the customers' perspective. 

For future work, we suggest employing other pre-trained language models and carrying out tests on more datasets. Also, utilizing a multilingual dataset for predicting online reviews' helpfulness can further showcase the true capabilities of multilingual language models and be helpful in building models for different languages simultaneously. We leave comprehensive multilingual data collection and model design to future research.

The code used in this paper is publicly available at~\url{https://github.com/AliBol/DSS-project}. The data used in this study is
publicly available at~\url{http://jmcauley.ucsd.edu/data/amazon/links.html}.

\bibliographystyle{splncs03}
\bibliography{references}

\end{document}